\documentclass{article}

\usepackage{microtype}
\usepackage{graphicx}
\usepackage{subfigure}
\usepackage{booktabs} 

\usepackage{hyperref}

\usepackage[accepted]{icml2024}

\usepackage{amsmath}
\usepackage{amssymb}
\usepackage{mathtools}
\usepackage{amsthm}

\usepackage{pgfplots}
\usepackage{placeins}

\usepackage{listings}
\lstdefinestyle{wraptt}{
  basicstyle=\ttfamily\small,
  breaklines=true,
  breakatwhitespace=false,
  columns=fullflexible,
  keepspaces=true,
  upquote=true
}

\usepackage[capitalize,noabbrev]{cleveref}

\theoremstyle{plain}

\theoremstyle{definition}

\theoremstyle{remark}

\usepackage[textsize=tiny]{todonotes}
\usepackage{multirow}
\usepackage{enumitem}

\DeclareTextSymbol{\textquotedbl}{OT1}{34}

\icmltitlerunning{LLM-AR: LLM-powered Automated Reasoning Framework}

\begin{document}

\twocolumn[
\icmltitle{LLM-AR: LLM-powered Automated Reasoning Framework}



\icmlsetsymbol{equal}{*}

\begin{icmlauthorlist}
\icmlauthor{Rick Chen}{oxford}
\icmlauthor{Joseph Ternasky}{vela}
\icmlauthor{Aaron Ontoyin Yin}{vela}
\icmlauthor{Xianling Mu}{oxford}
\icmlauthor{Fuat Alican}{vela}
\icmlauthor{Yigit Ihlamur}{vela}
\end{icmlauthorlist}

\icmlaffiliation{oxford}{Department of Mathematics, University of Oxford, Oxford, United Kingdom}
\icmlaffiliation{vela}{Vela Research, San Francisco, United States}

\icmlcorrespondingauthor{Rick Chen}{rick.chen@seh.ox.ac.uk}
\icmlcorrespondingauthor{Yigit Ihlamur}{yigit@vela.partners}

\icmlkeywords{Machine Learning, LLM, Automated Reasoning}

\vskip 0.3in
]



\printAffiliationsAndNotice{}  

\begin{abstract}
Large language models (LLMs) can already identify patterns and reason effectively, yet their variable accuracy hampers adoption in high-stakes decision-making applications. In this paper, we study this issue from a venture capital perspective by predicting idea-stage startup success based on founder traits. \textbf{(i)} To build a reliable prediction model, we introduce \textsc{LLM-AR}, a pipeline inspired by neural-symbolic systems that distils LLM-generated heuristics into probabilistic rules executed by the ProbLog automated-reasoning engine. 
\textbf{(ii)} An iterative policy-evolution loop incorporates association-rule mining to progressively refine the prediction rules. 

On unseen folds, \textsc{LLM-AR} achieves \(\mathbf{59.5\%}\) precision and \(\mathbf{8.7\%}\) recall, \(\mathbf{5.9\times}\) the random baseline precision, while exposing every decision path for human inspection. The framework is interpretable and tunable via hyperparameters, showing promise to extend into other domains.
\end{abstract}

\section{Introduction}
In the real world, high-stakes decision-making problems arise naturally in domains such as medicine, finance and law, characterized by their complexity and societal importance. Early-stage startup investment in venture capital (VC) is one such challenge. At the idea-stage, when the outcome is uncertain and information is limited, only about $1.9\%$ of startups eventually become outliers (defined in Section \ref{sec:dataset}). Identifying these startups with high precision is crucial for the efficient allocation of investment resources. With the emergence of large language models (LLMs), evaluating their reasoning abilities as potential VC analysts has become an increasingly compelling direction of research. 

Although LLMs exhibit strong reasoning capabilities across domains such as medical diagnosis \cite{SDBench} and VC \cite{chen2025vcbenchbenchmarkingllmsventure}, essential qualities like transparency and prediction stability are often undermined by their black-box nature. This motivates the integration of black-box models with traditional machine learning techniques, a strategy aimed at enhancing the explainability of LLM-based reasoning. Symbolic AI, in particular \textbf{automated reasoning systems} \cite{AutomatedReasoningFoundation}, excels in transparent and deterministic decision-making.

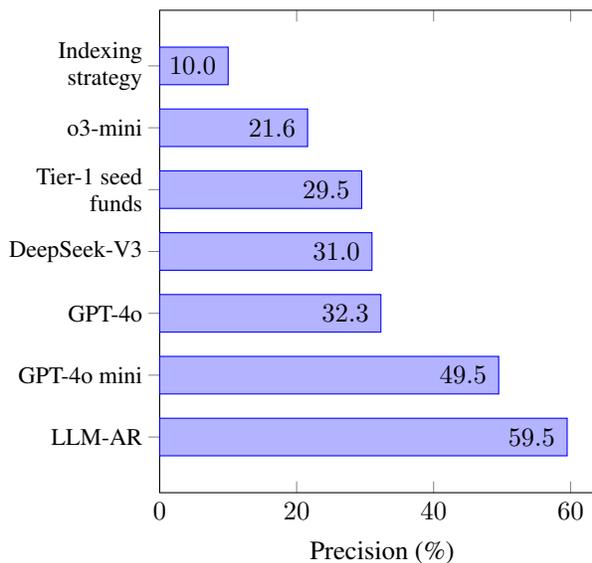
\begin{figure}[t!]
    \begin{tikzpicture}
        \begin{axis}[
            width=7.5cm, height=8cm,
            xbar,
            symbolic y coords={LLM-AR, GPT-4o mini, GPT-4o, DeepSeek-V3, {Tier-1 seed\\funds}, o3-mini,{Indexing\\strategy}},
            ytick=data,
            y tick label style={font=\small,  inner sep=2pt, align=right},
            nodes near coords,
            nodes near coords={\pgfmathprintnumber[fixed,precision=1,zerofill]{\pgfplotspointmeta}},
            nodes near coords align={horizontal}, 
            nodes near coords style={
                anchor=east,   
                xshift=-1pt,   
                text=black
                },
            xlabel={Precision (\%)},
            ylabel={},
            xmin=0, xmax=65,
            bar width=0.5cm,
            enlarge y limits=0.15,
            ]
            \addplot coordinates {(59.5,LLM-AR)(49.5,GPT-4o mini)(32.3,GPT-4o)(31.0,DeepSeek-V3)(29.5,{Tier-1 seed\\funds})(21.6,{o3-mini})(10.0,{Indexing\\strategy})};
        \end{axis}
    \end{tikzpicture}
    \vspace{-15px}
    \caption{Human-baseline precisions compared with average model precisions obtained by LLMs and LLM-AR on our dataset. Human-baseline precisions are scaled up linearly to reflect the inflation of success rate from the real-world (1.9\%) to our dataset (10\%).}
    \label{fig:modelcomparision}
\end{figure}

In this paper, we introduce \textbf{LLM-AR}, a two-stage decision-making model framework that combines the reasoning powers of LLMs with automated reasoning systems. The LLM generates policies consisting of prediction rules, which are subsequently processed by the automated reasoning component to yield deterministic predictions. In our implementation, each rule is associated with a probability score to address textual ambiguity. In addition, policies are trained through an iterative process, in which rules are added, deleted, or modified based on test statistics and insights gained from previous iterations.

We conduct a comprehensive model analysis in the context of startup success predictions. On a set of 6,000 founders, our model achieves a cross-validated precision of $\mathbf{5.9\times}$ the random success rate, surpassing both human and LLM baselines (Figure \ref{fig:modelcomparision}).

\textbf{Contributions. } Our model has the following key advantages:
\begin{enumerate}[noitemsep, topsep=0pt]
    \item \textbf{Explainable}: Policies are fully human-readable, enabling human experts to interpret and analyze the reasoning process of LLMs.
    \item \textbf{Modifiable}: Human experts can directly modify policies based on their own domain knowledge (``expert-in-the-loop").
    \item \textbf{Tunable}: By adjusting hyperparameters, one can trade off precision for recall based on needs, or vice versa.
    \item \textbf{High Precision}: Our continuous-scale outputs achieve higher precision than binary classification (see Figure~\ref{fig:modelcomparision}).
    \item \textbf{Generalizable}: The LLM-AR framework generalizes naturally to other high-stakes decision-making domains, such as medical diagnosis. 
\end{enumerate}

\section{Related Work}

\textbf{LLMs in VC. }Various LLM-based decision-making models have been proposed for VC applications. In the GPTree model \cite{gptree}, the LLM is prompted to generate a decision tree for startup predictions. After normalizing against the real-world success rate of $1.9\%$, the model achieved a cross-validated precision of $7.2\%$, demonstrating the potential of augmenting LLM-based reasoning with traditional machine learning techniques. The idea was later extended to random forest ensembles \cite{griffin25}.

Two studies that are more closely related to LLM-AR \cite{mu2025, preuveneers2025reasoningbasedaistartupevaluation} explored the generation of decision-making policies using LLMs, improving precision through techniques such as chain-of-thought prompting, in-context-learning and reinforcement learning.

\textbf{Automated Reasoning Systems. }Symbolic AI was prominent in the early development of AI, prior to the popularization of neural networks. A key subclass, {automated reasoning systems \cite{AutomatedReasoningFoundation}, are designed specifically to process logical statements, apply sound inference rules, and derive conclusions deterministically. Such systems underpin automated theorem provers and mathematical proof assistants \cite{IsabelleAR}. We explain our use case in Section~\ref{Section:AR}.

\textbf{Neural-Symbolic AI. }Neural-Symbolic AI \cite{NeuralSymbolicSurvey} is a more recent paradigm that integrates neural networks into symbolic AI systems. One of its core objectives is to combine the statistical learning capacity of neural networks with the interpretability and reasoning power of symbolic logic. Our LLM-AR framework is inspired by the influential NS-VQA system \cite{NSVQA}, which disentangles perception and reasoning for visual question answering. Visual patterns are recognized by convolutional neural networks, while answers are produced by executing deterministic symbolic programs generated from the questions.

\section{Dataset}\label{sec:dataset}
Our experiments are conducted on a curated and feature-engineered founder dataset consisting of 6,000 US-based founders. 600 (10\%) are labeled as successful. Each founder is paired with their startup in inspection, which was founded in or after 2010. A founder is considered successful if their startup was acquired or had an initial public offering (IPO) with a valuation exceeding \$500M, or raised more than \$500M in funding. Such founders are outliers in the population. On the other hand, the founder is labeled unsuccessful if their startup raised between \$100K and \$4M. Founder profiles were initially collected from LinkedIn and Crunchbase. 

\textbf{Numerical Features. }Textual profile information was converted into structured, numerical features suitable for quantitative analysis. For example, education attainment was mapped onto a normalized categorical scale, assigning a score of 3 to PhD holders and 1 to Bachelor's degree holders. Other directly extracted features include \texttt{number\_of\_work\_experience} and \texttt{vc\_experience}. In addition to explicit attributes, the dataset also attempted to encode implicit founder qualities, such as \texttt{perseverance} and \texttt{risk\_tolerance}. We used 52 features in total for model training and evaluation.

\textbf{Data Sampling and Anonymization. }For each founder, only data prior to the founding of their startup is included, simulating the lack of information of early-stage startups. Moreover, each profile is inherently anonymized through the textual-to-numerical conversion. This mitigates data contamination \cite{datacontam24}, a phenomenon in which an LLM bypasses the prediction task by directly recalling founders from its pre-training corpus.

\textbf{Dataset Splits. }We divide the dataset into four equal folds and perform cross-validation to ensure the robustness of results. We allocate two folds to training, one to validation and one to testing.

\section{Choice of Automated Reasoning System}\label{Section:AR}
In the LLM-AR framework, the choice of the automated reasoning system determines the nature of the model. For our purpose, PyDatalog \cite{pydatalog} was initially proposed for simplicity and effectiveness. PyDatalog is founded on the Prolog framework, a declarative logic programming language that uses first-order logic. This formalism allows the use of predicates (founder attributes), logical connectives (AND, OR, etc.) and implications. The following demonstrates a minimal working PyDatalog program in which a founder is predicted to be successful if they have worked at both Google and Apple:
\begin{lstlisting}[style=wraptt]
pyDatalog.create_terms('X, success,
    google, apple')

+google('Steve')
+apple('Steve')

success(X) <= google(X) & apple(X)

print(success(X))
\end{lstlisting}
To ensure correct generation of such programs based on LLM output, textual arguments need to be reliably and faithfully translated into symbolic logical statements. A key barrier to this conversion is textual ambiguity. Natural language arguments frequently contain vague or probabilistic quantifiers, such as ``most", ``usually", or ``strong". These expressions lie outside the expressiveness of first-order logic.

\subsection{ProbLog}
To resolve the issue of textual ambiguity, we propose introducing probabilities to model these ambiguities, and employing ProbLog  \cite{problog} as the automated reasoning system. In ProbLog, probabilities can be assigned to both logical implications and predicates, which in our context may be interpreted as ``weights". The following demonstrates an example program in ProbLog:
\begin{lstlisting}[style=wraptt]
0.7::education.
0.2::work_experience.

0.6::success :- education,work_experience.

query(success)
\end{lstlisting}
Now a query would calculate the probability of success or failure based on the probabilistic rules. As the number of rules increases, exact inference becomes computationally infeasible; ProbLog therefore implements an approximation sampling method \cite{problog}. Despite this stochastic approximation, the outcome remains reproducible and explainable because the underlying rules and their associated probabilities are fixed and transparent.

Finally, we model success and failure probabilities separately. This aims to improve precision by not only picking out candidates with positive indicators but also filtering out those who are likely to fail at the same time.

\section{Methodology}
\begin{figure*}[t]
    \centering
    \includegraphics[width=\textwidth]{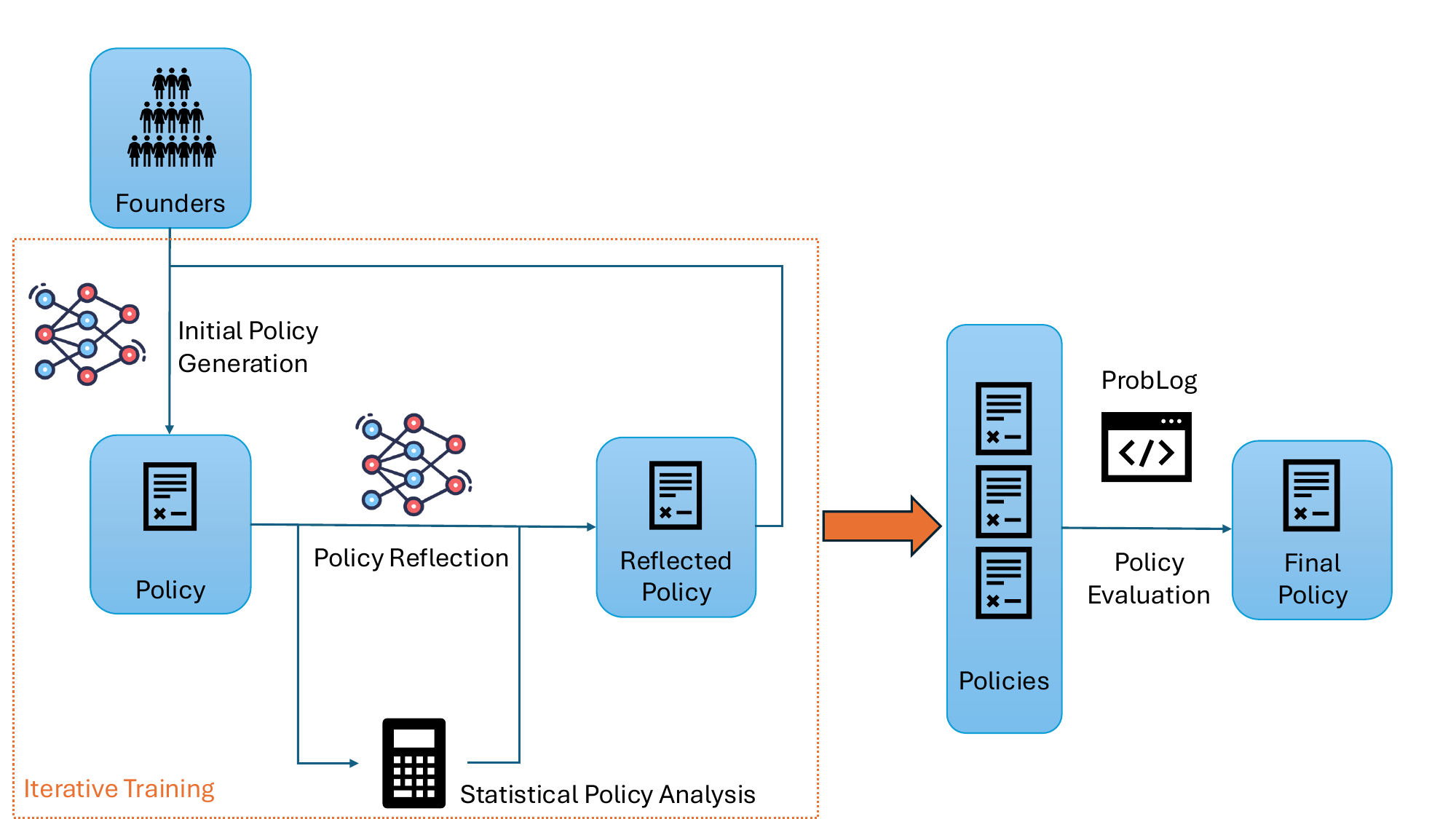}
    \caption{Our LLM-AR pipeline: the iterative training phase produces policies, which are then evaluated by the automated reasoning component. The iterative training phase involves repeated policy generation and reflection based on statistical policy analysis.}
    \label{fig:pipeline}
\end{figure*}
Our implementation of the LLM-AR framework is illustrated Figure \ref{fig:pipeline}. The model consists of four main conceptual components: (i) policy generation via LLM prompting; (ii) ProbLog policy evaluation; (iii) iterative policy evaluation and refinement; (iv) final policy selection and hyperparameter tuning.

At a high level, the LLM is first prompted to generate textual insights explaining why each founder succeeded or failed. These individual insights are aggregated and summarized into generalizable logical prediction rules, forming a policy. Through a translation mechanism, the resulting policy can be expressed in ProbLog syntax, which allows the computation of success and failure probabilities for unseen founders. As the model processes more founders, the prediction rules are iteratively refined based on both statistical analysis and LLM reflection.

\subsection{Iterative Training for Policy Generation}
The iterative training process encourages the LLM to add, delete, or modify any of the existing rules at each iteration. The training set is divided into small batches while maintaining the baseline prevalence of 10\%. For each batch, the following procedure is applied repeatedly (Figure \ref{fig:pipeline}):

\begin{enumerate}[noitemsep, topsep=0pt]
    \item Initial policy generation;
    \item Statistical policy analysis;
    \item Policy reflection.
\end{enumerate}

\subsubsection{Initial policy generation}
For each labeled founder in the batch, the LLM produces textual insight on why they succeeded or failed. Following \cite{gptree}, we employ the following prompt:

\begin{lstlisting}[style=wraptt]
"You are a VC anaylst trying to predict the success of a startup based on some information about the founders.

Given the following founder's background and startup idea:
    Founder Profile (LinkedIn followed by Crunchbase): {founder_profile}
This startup was eventually {success_text}.
Clearly explain the most important reasons why this startup {success_verb}. Try to use simple vocabulary" 
\end{lstlisting}

Once all founders in a batch are processed, the LLM is instructed to summarize the aggregated insights into a list of logical prediction rules of the form:
\begin{lstlisting}[style=wraptt]
IF condition_1 AND condition_2 
THEN success/failure.
\end{lstlisting}
At the same time, the LLM assigns a probability (confidence score) to each implication. Example outputs are provided in Appendix \ref{app:example_model_output}.

\subsubsection{Statistical policy analysis}
This step aims to provide statistical grounding for the LLM-generated policy.

\textbf{Rule calibration. }We apply association-rule mining \cite{agrawal1994fast}, a machine learning technique, to identify feature combinations that are statistically associated with success/failure. These combinations serve as hints for the LLM, ensuring that statistically significant relationships are reflected in subsequent policy updates.

\textbf{Probability calibration. }Probabilities associated with the prediction rules are calibrated using a dynamic sample of 1,000 observations from the training data. For each rule, we filter founders matching its preconditions and then compute the empirical success/failure proportion as an estimate of the true population proportion. If the filtered sample is too small, the rule is flagged with ``not enough samples".

\subsubsection{Policy reflection}
Finally, the LLM is asked to reflect on its initial policy based on the results of the statistical analysis \cite{Reflexion}. A new policy is produced and carried forward to the next iteration as guidance. The reflection prompt (see Appendix \ref{app:example_prompts} for an example) encourages the model to do as follows:

\begin{enumerate}[noitemsep, topsep=0pt]
    \item refine the probabilities;
    \item remove rules with insufficient statistical support, or rules with low predictive powers, statistically;
    \item incorporate new rules identified by association rule mining.
\end{enumerate}

\begin{table*}[ht]
\centering
\begin{tabular}{l|c c c|c c c}
\toprule
 & \multicolumn{3}{c}{Validation} & \multicolumn{3}{c}{Test} \\
\cmidrule(lr){2-4} \cmidrule(lr){5-7}
Fold & Precision (\%) & Recall (\%) & $F_{0.25}$ (\%)& Precision (\%) & Recall (\%)& $F_{0.25}$ (\%)\\
\midrule

\multirow{2}{*}{1} & 64.3 & 9.0 & 47.2 &	58.3 &7.0	&40.7\\
&71.4&	5.0&	40.1&	66.7&	9.0&	\textbf{48.4}\\
\multirow{2}{*}{2}& 80.0&	8.0&	52.3&	52.4&	\textbf{16.5}&	46.5\\
&45.0&	9.0&	36.4& 59.0&	11.5&	47.5\\
\multirow{2}{*}{3}& 59.3&	16.0&	51.2&	65.4&	8.5&	46.9\\
 &60.0&	12.0&	48.6&	62.1&	9.0&	46.1\\
\multirow{2}{*}{4}& 72.7&	8.0&	49.3&	\textbf{83.3}&	2.5&	28.7\\
&50.0&	6.0&	34.9&	56.3&	9.0&	43.0\\
\multirow{2}{*}{5}& 66.7&	8.0&	46.6&	57.9&	5.5&	37.1\\
 &77.3&	\textbf{17.0}&	\textbf{64.0}&	47.1&	12.0&	40.2\\
\multirow{2}{*}{6}& 58.8&	10.0&	45.7&	50.0&	8.5&	38.8\\
&\textbf{88.9}&	8.0&	55.7&	55.6&	5.0&	34.9\\

\midrule
Avg. & 66.2&	9.7&	47.7&	\textbf{59.5}&	\textbf{8.7}&	\textbf{41.6}\\

\bottomrule
\end{tabular}
\caption{Model performance metrics for our LLM-AR model, across 12 different partitions of the dataset.}
\label{tab:results}
\end{table*}

\subsection{Problog Prediction using Policies}
For a given policy and founder, a prediction is obtained through the following deterministic pipeline:

\begin{enumerate}[noitemsep, topsep=0pt]
    \item Convert each prediction rule into ProbLog format.
    \item Convert founder traits into ProbLog format, assigning higher probabilities to stronger characteristics.
    \item Execute two ProbLog queries to compute the success and failure probabilities.
    \item Classify the founder as successful if their success probability exceeds the success threshold and their failure probability is below the failure threshold, where the two thresholds are tunable hyperparameters.
\end{enumerate}

We optimize these thresholds based on the $F_{0.25}$-score (Eq.~\ref{eq:F0.25}), which assigns four times more weight to precision than recall. This reflects the emphasis on pushing for high precision and minimizing false positives in the VC context.

\begin{equation}\label{eq:F0.25}
  F_{0.25} = (1 + 0.25^2) \cdot \frac{\text{Precision} \cdot \text{Recall}}{(0.25^2 \cdot \text{Precision}) + \text{Recall}}  
\end{equation}

To determine the optimal success and failure thresholds for a given policy, we run ProbLog predictions on the validation set. This is repeated for numerous threshold pairs, and the combination that produces the best validation $F_{0.25}$-score is retained for the final testing.

\subsection{Training-time Policy Evaluation}
Because the iterative training procedure does not guarantee monotone improvement over time, we perform a training-time policy evaluation every five iterations. The policies of the five most recent iterations are tested on a held-out subset of the validation set. The performance metrics (in particular the $F_{0.25}$-scores) are sequenced and analyzed by an LLM evaluator. The evaluator is instructed to examine how rules and performance evolved over time, identify which modifications may have improved/degraded performance, and generate an updated policy accordingly. This helps prevent the policy from developing on a declining path.

\subsection{Final Policy and Hyperparameter Selection}
There is one policy produced at each iteration until the final one. Instead of selecting the policy from the final iteration by default, we test policies from multiple selected iterations on the remaining validation set. The policy achieving the highest $F_{0.25}$-score is selected as the output policy, together with its corresponding (optimized) success and failure thresholds.

\section{Results}
\subsection{Cross-Validation}
To ensure the robustness of the results, we divided the dataset into four equal parts and performed 4-fold cross-validation. This included all combinations of choosing two folds for training (6 ways), one for validation (2 ways), and the remaining for testing, yielding 12 partitions in total.

\subsection{LLM-AR}
The cross-validated results of our LLM-AR implementation are summarized in Table \ref{tab:results}. The primary LLM used was \textbf{Deepseek-V3}. Due to computational constraints, only ten iterations of the iterative training process were executed, and only four candidate policies were tested.  Our LLM-AR implementation achieved an average precision of \textbf{59.5\%} and an average recall of \textbf{8.7\%}, representing a \textbf{5.9$\times$} improvement over the baseline precision. 

\subsection{Ablation Analysis}

\textbf{Without iterative training.} By testing only the initially generated policies, we observe a substantial drop in $F_{0.25}$-score. Table \ref{table:iteration_0_results} summarizes the average performance metrics across all 12 partitions of the dataset. This confirms the effectiveness of iterative training.

\begin{table}[h]
\centering
\begin{tabular}{l c c c c}
\toprule
Model &  Precision & Recall & $F_{0.25}$\\
\midrule
With iterative training &  \textbf{59.5\%} & 8.7\% & \textbf{41.6\%} \\
without iterative training  & 36.1\% & \textbf{14.2\%} & 32.7\% \\
	
\bottomrule
\end{tabular}
\caption{Average policy performance with/without iterative-training.}
\label{table:iteration_0_results}
\end{table}

\textbf{Without ProbLog. }We also experimented with replacing ProbLog with an LLM, which is prompted with a selected iteratively-trained policy. There is again a drop in average $F_{0.25}$-score (Table \ref{table:without_problog_results}), showing that ProbLog offers not only transparent but also more effective probabilistic reasoning.

\begin{table}[h]
\centering
\begin{tabular}{l c c c c}
\toprule
Model & Precision & Recall & $F_{0.25}$\\
\midrule
ProbLog & \textbf{59.5\%} & 8.7\% & \textbf{41.6\%} \\
GPT-4o mini with policy  & 46.2\% & \textbf{12.0\%}& 39.5\% \\
\bottomrule
\end{tabular}
\caption{Average policy performance when ProbLog is replaced with GPT-4o mini.}
\label{table:without_problog_results}
\end{table}

\subsection{Vanilla LLMs}
To establish baseline LLM results, we divided all 6,000 founders into six folds. Each LLM was prompted to classify founders as successful or unsuccessful using a specifically designed founder-profile format; the numerical features are converted back into textual format by reversing the original textual-to-numerical encoding. Since our LLM-AR pipeline relies heavily on these numerical features, this profile format is appropriate as an ablation analysis. Furthermore, data contamination \cite{datacontam24} is mitigated as no explicit identifiers can be recovered through this conversion (as discussed in Section \ref{sec:dataset}).

The average results are presented in Table \ref{table:vanilla_LLM_textual}, with fold-specific results included in Appendix \ref{app:vanilla_llm_results}. Among the four models tested, GPT-4o mini achieved the highest precision (49.5\%).

\begin{table}[h]
\centering
\begin{tabular}{l c c c}
\toprule
Model & Precision (\%) & Recall (\%) & $F_{0.25}$ (\%) \\
\midrule
GPT-4o mini & \textbf{49.5} & 8.0 & \textbf{37.8} \\
GPT-4o     & 32.3 & 18.7 & 30.9 \\
DeepSeek-V3 & 31.0 & 15.2 & 29.1 \\
o3-mini    & 21.6 & \textbf{31.3} & 22.0 \\
\bottomrule
\end{tabular}
\caption{Average vanilla LLM baselines across six folds.}
\label{table:vanilla_LLM_textual}
\end{table}

\subsection{Model Tunability}
Looking beyond our use case, there are scenarios such as medical diagnosis in which high-recall models are preferred. One could then select a model with appropriate weightings on precision and recall, by optimizing the $F_\beta$-score for a suitable $\beta$. To demonstrate this, we selected a representative policy and determined the threshold pairs that maximized the $F_{\beta}$-score for different $\beta$. This yielded a broad spectrum of model behaviors (Table \ref{table:different_F-scores}) without any retraining.

Furthermore, the iterative training process itself depends on $\beta$, because training-time policy evaluation examines the $F_{\beta}$-score. Consequently, in practical generalizations, one can expect a higher performance than that reported in Table \ref{table:different_F-scores}, since training-time policy evaluation actively guides policy refinement towards higher $F_{\beta}$-scores for the chosen $\beta$.

\begin{table}[h]
\centering
\begin{tabular}{l c c c c}
\toprule
$\beta$ & Precision (\%) & Recall (\%) & $F_{\beta}$ (\%)\\
\midrule
4 & 12.5 &  92.0 &  66.9 \\
2   &  15.9& 72.0 & 42.2 \\
1   & 30.6 & 36.0 & 33.1 \\
0.5    & 43.5 & 20.0 & 35.2 \\
0.25     & 59.3 & 8.0 & 43.0 \\
0.125      & 100.0 &  2.0 & 57.0 \\

\bottomrule
\end{tabular}
\caption{By optimizing the $F_\beta$-score for different $\beta$, a range of models with different predictive powers can be obtained.}
\label{table:different_F-scores}
\end{table}

\section{Discussion}
After cross-validation, our LLM-AR model achieved a precision of 59.5\% and a recall of 8.7\%, exceeding the random success rate by 5.9$\times$. Although recall was relatively low, we valued precision over recall to minimize the allocation of investment resources to false positives; one could also optimize for the $F_1$-score to guarantee higher recall. The drop in model performance from validation to test sets was expected, as optimal success and failure thresholds vary across data partitions. Nevertheless, most of the precision was retained, suggesting that thresholds learned from one partition generalize reasonably well to others.

\subsection{Limitations}

\textbf{Prevalence shift. }The baseline prevalence of our feature-engineered dataset is 10\%, inflated from the real-world figure of 1.9\%. Predictive performance does not necessarily scale linearly, so the real-world performance of LLM-AR should be interpreted cautiously.

\textbf{Constraint on features. }Throughout our experiments, LLMs were restricted to using only founder traits available in the fixed feature-engineered dataset, since statistical analysis is reliant on already existing numerical data.

\textbf{Runtime limitations. }With the ProbLog library in Python, we encountered difficulties in scaling policy size and reducing policy evaluation time. Substantial optimization appears possible, as the official ProbLog web implementation\footnote{https://dtai.cs.kuleuven.be/problog/editor.html} performs considerably faster.

\textbf{Limited control over policy convergence. }Beyond the training-time policy evaluations, there is limited control over the progressive improvement of the policies across iterations. This was circumvented by testing multiple policies instead of the final one alone.

\subsection{Future Work}
\textbf{Component-level interpretability. }The presence of multiple interacting components in our model makes it difficult to analyze comprehensively how each component contributes. A systematic component-level analysis would provide valuable insights and guide future improvements.

\textbf{Scaling of probabilities. }Given the inherently low baseline success rate, rule probabilities often cluster within a narrow range. We rescaled the probabilities prior to ProbLog evaluation to further separate high-probability rules from low-probability ones, which encourages high-precision models. Future work should explore alternative rescaling or normalization methods.

\textbf{LLM-powered feature selection. }Allowing LLMs to propose new features could reveal more natural and powerful prediction rules. One possible approach is to continuously collect and encode data based on LLM suggestions.

\textbf{Alternative statistical policy analysis methods. }While our LLM-AR implementation relied on statistical calculations\footnote{It should also be noted that Problog itself offers the option to learn probabilities from data, which we have not exploited.} and association-rule mining, other machine learning techniques such as Random Forest and Bayesian Network are worth exploring. The Bayesian network approach in particular could enable the encoding of multi-step reasoning processes; intermediate implications such as ``professional athlete implies perseverance" may be expressed, rather than requiring success or failure as conclusions of all logical implications. 

\textbf{Alternative LLM-AR implementations. }LLM-AR provides a broad framework for decision-making models. Component-wise adaptations, in particular the use of different automated reasoning or Symbolic AI systems, can produce a diverse family of models. One could also experiment with different LLMs to maximize performance.

\textbf{Experimenting in other domains. }Future work should explore adaptations of the LLM-AR framework in other high-stakes decision-making domains, in particular medical diagnosis.

\section{Conclusion}
In this paper, we introduced LLM-AR, an explainable decision-making framework combining the reasoning powers of LLMs and automated reasoning systems. With this architecture, the LLMs generate human-interpretable policies consisting of prediction rules, which are then processed by an automated reasoning module to produce transparent and deterministic predictions. For our implementation, we adopted ProbLog to introduce probabilistic confidence scores to both the prediction rules and the founder traits. We also proposed an iterative training process, in which policies are progressively refined through updated knowledge, statistical calibrations, and self-reflections. 

Our LLM-AR model achieved a cross-validated precision of 59.5\%, representing a 5.9$\times$ improvement over the random success rate. Furthermore, our model has the advantages of being explainable, tunable, modifiable, and generalizable. Looking ahead, we anticipate the adaptation of this architecture to other high-stakes decision-making domains, such as medicine, finance and law.


\bibliography{main}
\bibliographystyle{icml2024}

\appendix
\section*{Appendix}
\section{Example Model Outputs}\label{app:example_model_output}

\subsection{Example Founder Insight}

\begin{lstlisting}[style=wraptt]
Here are the key reasons why this startup succeeded, based on the founder's background:

1. **Deep Industry Experience** - The founder worked in financial services and tech for over 20 years before starting the company. This means they knew the industry well, understood problems, and had the right connections.

2. **Multiple Founder Roles** - They had already founded or helped start at least three other companies (Fluid Trade, Altenex, Worldstreet). This shows they know how to build a business from scratch.

3. **Leadership Skills** - They held executive roles like SVP, CEO, and Managing Partner, proving they could lead teams and make big decisions.

4. **Diverse Background** - They worked in finance, renewable energy, real estate, and software. This mix likely helped them see unique opportunities others missed.

5. **Strong Network** - Working at multiple companies means they built relationships with investors, customers, and talent-key for startup success.

6. **Persistence** - Their career shows long-term commitment (e.g., 6 years at Altenex, 2+ years at other startups). Startups take time, and this founder had patience.

In short: They had the right experience, leadership skills, and network to make the startup work. Past failures or lessons from other ventures likely helped too.
\end{lstlisting}

\subsection{Example Policy}

\begin{lstlisting}[style=wraptt]
Success rules:
num_acquisitions AND career_growth,0.40
perseverance AND vision,0.32

Failure rules:
not_career_growth AND not_num_acquisitions,0.96
not_education_level AND not_education_institution,0.89
\end{lstlisting}

\subsection{Example Rule Reflection}
\begin{lstlisting}[style=wraptt]
After analyzing the data probabilities compared to my initial intuition, here's the modified set of rules with adjustments:

1. Success rules where probabilities were reasonably close (within ~0.05 difference) were kept with adjusted probabilities
2. Success rules with significantly lower probabilities than expected were either:
   - Kept with lowered probabilities if still above random chance (0.1)
   - Removed if below random chance threshold
3. Failure rules generally matched well and were kept as is
4. Rules with "not enough samples" were removed due to insufficient evidence
5. Incorporated the high-probability success rule about CEO experience and acquisitions since it aligns with existing rules

Modified rules:

(...)

Key changes made:
(...)

The modified rules now better reflect the actual patterns found in the data while maintaining the original rule structure.
\end{lstlisting}

\section{Example Rule Reflection Prompt}\label{app:example_prompts}
\begin{lstlisting}[style=wraptt]
You are a venture capital analyst evaluating startup success patterns.

You previously created intuitive logical rules about how founder attributes relate to success.

Now, compare those intuitive rules to empirical data, assess consistency, and revise the rules to better reflect observed probabilities.

Below are your original intuitive rules:

{logical_statements}

Here are the same rules with probabilities estimated from real data:

{calibrated_statements}

You are also optionally given:
- A few **high-probability success rules** discovered from the data:
  {success_rule_hints}

- One **high-probability failure rule** discovered from the data:
  {failure_rule_hints}

You may incorporate these hints only if they fit coherently within your revised logic.
---

### **Reasoning & Update Instructions**
1. **Compare** your intuitive probabilities with the data-calibrated ones.
2. If they differ, reason about *why*:  
   - Were your intuitions biased or based on rare/outlier cases?  
   - Is there insufficient data ("not enough samples")? If so, treat that rule as unreliable.
3. **Prune or adjust rules** as follows:
   - Remove rules with success probability < 0.1 (too low relative to random baseline 0.1).  
   - Remove rules with failure probability < 0.9 (too low relative to random baseline 0.9).  
   - Modify probabilities or logic as needed to align with the data.
4. You may **delete or modify** existing rules, but **must not create entirely new ones**.
5. Ensure all features follow the required naming format.

---

### **Output Requirements**
Return only your **final, modified rules** in exactly the same format as the original rules.  
Double-check that all deletions and adjustments are correctly applied.
Do not include explanations or commentary-only the updated rules.
"""
\end{lstlisting}
\section{Fold-specific Vanilla LLMs Result Tables}\label{app:vanilla_llm_results}

\begin{table}[H]
\centering
\begin{tabular}{l c c}
\toprule
Fold & Precision (\%) & Recall (\%) \\
\midrule
1	&57.1&	8.0 \\
2	&62.5	&10.0\\
3	&46.7	&7.0\\
4	&45.0	&9.0\\
5	&46.7	&7.0\\
6	&38.9	&7.0\\
\midrule
Average&	49.5	&8.0\\
\bottomrule

\end{tabular}
\caption{Vanilla GPT-4o-mini results}
\label{table:GPT-4o_mini_textual}
\end{table}

\begin{table}[H]
\centering
\begin{tabular}{l c c}
\toprule
Fold & Precision (\%) & Recall (\%) \\
\midrule
1&	30.9	&17.0\\
2	&38.0	&19.0\\
3	&34.4	&22.0\\
4	&29.5	&18.0\\
5	&28.1	&18.0\\
6	&32.7	&18.0\\

\midrule
Average&32.3	&18.7\\
\bottomrule

\end{tabular}
\caption{Vanilla GPT-4o results}
\label{table:GPT-4o_textual}
\end{table}

\begin{table}[H]
\centering
\begin{tabular}{l c c}
\toprule
Fold & Precision (\%)& Recall (\%) \\
\midrule
1	&26.8&	11.0 \\
2	&38.6	&17.0\\
3	&31.8&	14.0\\
4	&27.9	&17.0\\
5	&31.3	&15.0\\
6	&29.3	&17.0\\

\midrule
Average	&31.0	&15.2\\
\bottomrule

\end{tabular}
\caption{Vanilla DeepSeek-V3 results}
\label{table:DeepSeek_textual}
\end{table}

\begin{table}[H]
\centering
\begin{tabular}{l c c}
\toprule
Fold & Precision (\%)& Recall (\%)\\
\midrule
1&	20.1	&28.0\\
2	&23.8	&35.0\\
3	&23.7	&36.0\\
4	&19.9	&29.0\\
5	&20.4	&29.0\\
6	&21.8	&31.0\\

\midrule
Average&21.6	&31.3\\
\bottomrule

\end{tabular}
\caption{Vanilla o3-mini results}
\label{table:o3mini_textual}
\end{table}

\end{document}